\documentclass[11pt]{article}

\usepackage[preprint]{acl}

\usepackage{times}
\usepackage{latexsym}

\usepackage[T1]{fontenc}
\usepackage[utf8]{inputenc}

\usepackage{microtype}

\usepackage{inconsolata}

\usepackage{graphicx}
\usepackage{float}
\usepackage{amsmath} 
\usepackage{listings}
\usepackage{xcolor}

\usepackage{listings}
\usepackage{xcolor}

\lstdefinelanguage{JSON}{
  basicstyle=\ttfamily\small,
  numbers=left,
  numberstyle=\tiny\color{gray},
  stepnumber=1,
  numbersep=8pt,
  showstringspaces=false,
  breaklines=true,
  frame=single,
  backgroundcolor=\color{white},
  literate=
   *{:}{{\textcolor{black}{:}}}2
    {,}{{\textcolor{black}{,}}}2
    {\{}{{\textcolor{black}{\{}}}2
    {\}}{{\textcolor{black}{\}}}}2
    {[}{{\textcolor{black}{[}}}2
    {]}{{\textcolor{black}{]}}}2,
  stringstyle=\color{blue},
  morestring=[b]"
}

\title{
From Brute Force to Semantic Insight: \\ Performance-Guided Data Transformation Design with LLMs
}

  \author{Usha Shrestha,\space\space\space Dmitry Ignatov,\space\space\space Radu Timofte \\
       Computer Vision Lab, CAIDAS \& IFI, University of W\"urzburg, Germany}

\begin{document}
\maketitle

\begin{abstract}
   Large language models (LLMs) have achieved notable performance in code synthesis; however, data-aware augmentation remains a limiting factor, handled via heuristic design or brute-force approaches. We introduce a performance-aware, closed-loop solution in the NNGPT ecosystem of projects that enables LLMs to autonomously engineer optimal transformations by internalizing empirical performance cues. We fine-tune LLMs with Low-Rank Adaptation on a novel repository of 6,000+ empirically evaluated PyTorch augmentation functions, each annotated solely by downstream model accuracy. Training uses pairwise performance ordering (better–worse transformations), enabling alignment through empirical feedback without reinforcement learning, reward models, or symbolic objectives. This reduces the need for exhaustive search with fewer evaluated candidates than brute-force discovery while maintaining competitive peak accuracy and shifting generation from random synthesis to task-aligned design. Ablation studies show that structured Chain-of-Thought prompting introduces syntactic noise and degrades performance, whereas direct prompting ensures stable optimization in performance-critical code tasks. Qualitative and quantitative analyses demonstrate that the model internalizes semantic performance cues rather than memorizing syntax. These results show that LLMs can exhibit task-level reasoning through non-textual feedback loops, bypassing explicit symbolic rewards.
\end{abstract}

\section{Introduction}

A neural network’s performance depends on both its architecture and how the data is preprocessed. Effective data preprocessing and augmentation are essential for model generalization and convergence. Numerous studies have investigated optimal strategies for data transformations and the development of novel methods. For instance, AutoAugment \citep{cubuk2019autoaugmentlearningaugmentationpolicies} applies reinforcement learning to identify effective augmentation policies, while Meta Learning \citep{BILALLI2018101} approach uses a predictive meta-model to suggest data transformations for a specific classification algorithm. As generative artificial intelligence becomes increasingly prevalent, recent studies have begun to explore code generation using large language models (LLMs), leveraging their generative capabilities to propose and assess complex, data-aware solutions.

The NNGPT framework \citep{ABrain.HPGPT,ABrain.NNGPT} previously established a methodology for synthesizing neural network architectures using LLMs. Extending this work, we automate the generation and evaluation of data augmentation functions within the NNGPT ecosystem. This study also addresses the limited diversity of data augmentations in the LEMUR dataset, which comprises a broad range of high-capacity and edge-optimized neural network models \citep{ABrain.NN-Dataset,ABrain.LEMUR2,ABrain.NN-Lite} and serves as the knowledge base for the NNGPT. Motivated by recent advances in LLM applications across multiple domains \citep{Gado2025llm,Rupani2025llm,ABrain.NN-RAG} and prior NNGPT experiments \citep{ABrain.NN-Captioning_2025,ABrain.Prompt,ABrain.NNGPT-Fractal,ABrain.Architect}, we curate a diverse set of PyTorch data transformation functions and systematically evaluate their effects, producing performance-annotated metadata that links each code snippet to its impact on model training. This metadata is used to fine-tune the language model, enhancing its understanding of data variability and performance effects. We further implement a system that generates PyTorch data transformation functions and iteratively refines the generator through supervised fine-tuning.

\section{Related Works}

The automated generation of data augmentations \citep{yang2023survey} has evolved significantly. The process began with AutoAugment \citep{cubuk2019autoaugmentlearningaugmentationpolicies}, which uses reinforcement learning to search a fixed list of operations. This approach defines a discrete search space of 14 to 16 standard image processing methods, such as Rotate, ShearX, and ShearY. In this approach, generation refers to finding an optimal policy composed of sub-policies that specify two sequential transformation operations, the probability of applying each operation, and the magnitude of each operation. This method transforms the problem into a large-scale discrete search challenge, with AutoAugment's search space containing approximately $10^{32}$ possible policies.

The high computational cost of automated search in AutoAugment led to approaches that simplified the generation process. RandAugment \citep{cubuk2019randaugmentpracticalautomateddata} demonstrated that a complex search algorithm is unnecessary. Instead, it automated generation by reducing the search space to two interpretable hyperparameters: N (the number of transformations to apply) and M (a single, global magnitude for all transformations). RandAugment randomly samples N transformations from a predefined list and applies them with magnitude M. This generation method matched the performance of AutoAugment, indicating that the diversity of the transformation space is more critical than the complexity of the generation algorithm.

A key limitation of both AutoAugment and RandAugment is that they generate a policy that is applied to every image. However, a policy that is good for one image may be harmful to another \citep{Aboudeshish2025augmentation}. This led to the development of automated generation frameworks that are instance-specific and creates a unique augmentation policy for each individual image \citep{minh2021automatedimagedatapreprocessing}. For each image, a Deep Q-Network (DQN) iteratively generates a policy by selecting an action from a list of transformations or a "Stop" action. This process generates a unique, optimal chain of transformations for every sample in the dataset. This is a far more granular method of "generating a large number of transforms," as it generates one policy per instance rather than one policy per dataset.

Later research aimed to generate more effective and diverse transforms by expanding the space of possible transformations \citep{Mumuni_2025}. This expansion of the generation space occurred in two ways: through learned transformations and generative models. This approach enables the model to learn the transformation function itself. For example, Spatial Transformer Networks (STNs) \citep{Mumuni_2025} can be integrated into a model to learn optimal affine transformations directly from the data.

The latest paradigm in automated generation utilizes Large Language Models (LLMs) as the primary generation engine (\citealp{MUMUNI2025113}, \citealp{ding2024dataaugmentationusinglarge}). Recent approaches to adapting LLMs for specific tasks have shifted toward Instruction Tuning and Supervised Fine-Tuning (SFT) (\citealp{parthasarathy2024ultimateguidefinetuningllms}, \citealp{10.5555/3722577.3722647}). \citet{ouyang2022traininglanguagemodelsfollow} showed that fine-tuning models on human-written instructions aligns them more closely with user intent than simply increasing model size. This method is also effective in specialized domains, such as Python programming \citep{bai2022traininghelpfulharmlessassistant}. In code generation, where models must address complex and functional requirements beyond basic text completion, such alignment is crucial. \citet{chen2021evaluatinglargelanguagemodels} further confirmed this by demonstrating that optimizing general-purpose LLMs on code corpora enhances performance on functional correctness benchmarks.

Recent research in LLM adaptation also highlights the significance of data quality and training strategies over sheer data quantity. \citet{longpre2023flancollectiondesigningdata} identified that task balancing and enriching training data, such as by inverting input-output pairs, are essential for improving generalization. Their results indicate that combining zero-shot, few-shot, and Chain-of-Thought (CoT) data during fine-tuning leads to better performance across evaluation settings. 

The structure of input prompts and the training data have a significant impact on the quality of generated outputs, even though SFT updates model weights. A systematic survey by \citet{sahoo2025systematicsurveypromptengineering} categorizes advanced prompting techniques, including Chain-of-Thought (CoT) and decomposed prompting, which are essential for guiding models through multi-step logical tasks such as complex data augmentation. Building on these findings, \citet{kojima2023largelanguagemodelszeroshot} demonstrated that LLMs function as effective "zero-shot reasoners" and can perform complex task by simply appending the prompt "Let's think step by step." This approach enables the creation of reasoning-dense training data without the need for manual annotation.

In the field of code generation, AceCoder \citep{li2023acecoderutilizingexistingcode} addresses the challenge of requirement understanding through a guided code generation mechanism. This approach prompts the model to produce intermediate outputs, such as test cases or clarifications, prior to generating the final code. It instructs the model on "what to write" before determining "how to write it." Furthermore, LAIL (LLM-Aware In-Context Learning) \cite{li2023largelanguagemodelawareincontext} was introduced to ensure that high-quality examples are utilized during training or inference. This method filters training data based on the model's preferences, rather than relying on heuristic text similarity metrics, by employing a teacher LLM to estimate the likelihood that a given example will facilitate ground truth generation. These methods highlight a shift toward employing LLMs as active participants in code construction and quality assurance workflows, rather than just as final predictors.
	
\section{Methodology}
\label{sec:Methodology}
\subsection{LLM based code generation}
   We used the Olympic Coder 7B \citep{olympiccoder7b} model, an open-source AI model developed by Hugging Face. It is specifically designed to address complex olympiad-level programming problems and is fine-tuned on the CodeForces-CoTs \citep{penedo2025codeforces} dataset. We carried out code generation through various prompting approaches \citep{schulhoff2025promptreportsystematicsurvey, sahoo2025systematicsurveypromptengineering}, including Zero-shot prompting, Role prompting, Constraint prompting, and Chain-of-thought prompting methods. However, we noticed issues such as identical transform functions and syntax errors in the generated output.

\subsection{Brute Force Approach}
    After initial approaches yielded limited improvements, we adopted a manual method. First, we designed a system to automatically generate image transformation functions using the PyTorch  \citep{NEURIPS2019_9015} torchvision \citep{torchvision2016} package. The available transforms were organized into a dictionary. Given two parameters, the total number of files and the number of augmentations per file, we generated a number of transformation scripts. Each file contained one, two, or three selected transforms from the dictionary, in addition to fixed transforms: resize$(64,64)$, to tensor, and normalization. The generator permuted and cycled through various transform combinations to generate the files, assigning random parameters to each selected transform function. In total, 6,000 transform files were generated and evaluated, 2,000 for each case of using one, two, or three variable transforms. 
    
    \begin{figure}[h]
		\includegraphics[width=1.0\linewidth]{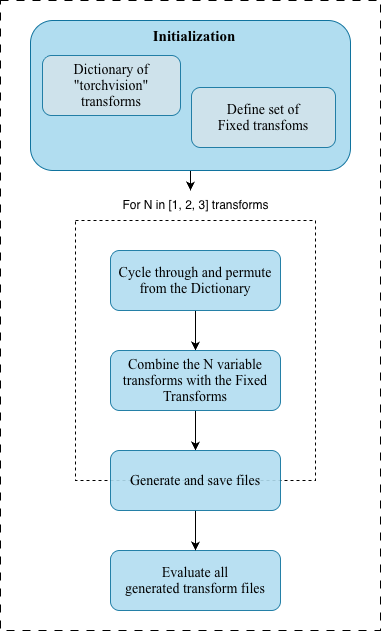}
		\caption{Brute-force data transformation generation and evaluation pipeline for constructing an LLM fine-tuning dataset. Image transformation functions are automatically generated, evaluated under a fixed training configuration, and stored with their corresponding accuracy, yielding a performance-labeled dataset used in subsequent LLM fine-tuning.}
		\label{rmse:ratio:train}	
	\end{figure}

\subsection{Fine Tuning LLM}

We employed an iterative instruction fine tuning approach that alternates between generating data transformations, evaluating their performance, and using the resulting metadata to refine the language model. To enable efficient adaptation without the computational overhead of full-parameter fine-tuning, we employed Low-Rank Adaptation (LoRA) \citep{hu2021loralowrankadaptationlarge} configured as shown in Listing 1 along with other hyperparameters. We utilized the set of generated image transforms and their evaluations obtained from the brute-force technique. 

\begin{multline*}
    \mathrm{LLM}_{\theta_t} 
    \xrightarrow{\text{Generate}} \{T_1, \dots, T_n\} \\
    \xrightarrow{\text{Evaluate}} \{(T_i, \mathrm{Acc}_i)\} 
    \xrightarrow{\text{Filter}} \mathcal{D}^{(t)} \\
    \xrightarrow{\text{Fine-Tune}} \mathrm{LLM}_{\theta_{t+1}}
\end{multline*}

First, a prompt is constructed using a few-shot strategy from a given prompt template by randomly selecting seed transforms${(T_i)}$ from the training data${\mathcal{D}^{(t)}}$. The LLM$(\mathrm{LLM}_{\theta_t})$ then generated several transforms that utilize common patterns from the references to optimize for specific task in each iteration${(t)}$. Each generated transform was evaluated using the same hyperparameters as the brute-force method. Generated data is filtered to identify better examples, specifically looking for instances where the generated transform improved upon the baseline accuracy. A dataset is constructed by iterating through each transform ("A") and searching for another transform as an ‘add-on’ transform ("B") with a higher accuracy to generate training pairings in which B outperforms A. This collection of "B better than A" pairs is formatted into instruction-tuning pairs using another prompt template that serves as the fine-tuning dataset. 

\noindent\begin{minipage}{\linewidth}
\begin{lstlisting}[language=Python, caption={ Hyperparameter configuration for LoRA fine-tuning and generation.}, label={lst:hyperparams}]
hyperparameters = {
    # LoRA Adapter Configuration

    # Rank of update matrices
    "r": 32,  
    # Scaling factor
    "lora_alpha": 32, 
    "lora_dropout": 0.05,
    "bias": "none",
    # Adapters applied to all attention projections
    "target_modules": [        
        "q_proj", "k_proj", 
        "v_proj", "o_proj"
    ],

    # Optimization Strategy
    "optimizer": "paged_adamw_8bit",  
    "learning_rate": 1.5e-4,
    "lr_scheduler_type": "cosine",
    "warmup_ratio": 0.05,
    # Epochs per fine-tuning iteration
    "num_train_epochs": 3,            

    # Batch Size 
    "per_device_train_batch_size": 1,
    "gradient_accumulation_steps": 8,
    "effective_batch_size": 8,       

    # Generation & Sampling
    # Controls diversity
    "temperature": 0.8, 
    # Nucleus sampling
    "top_p": 0.9,                     
    "top_k": 70,
    "max_new_tokens": 16 * 1024 
}
\end{lstlisting}
\end{minipage}

\section{Experiments and Results}

All data transformation functions were evaluated for the image classification task using a ResNet \citep{7780459} model on the CIFAR-10 \citep{krizhevsky2009learning} dataset. The model was trained for 1 epoch with a batch size of 64, a learning rate of 0.01, a momentum of 0.9, and a dropout rate of 0.2, due to resource and time constraints. All experiments, including the brute-force search and iterative fine-tuning loops, were conducted on a local workstation with a single NVIDIA GeForce RTX 4090 GPU (24 GB VRAM) which demonstrates the accessibility of our method for researchers with limited computational resources.

The Constraint method achieved the highest performance among LLM-based generation techniques. In this approach, the LLM was directed to modify a specified transform. Of the LLM generated transforms, approx. 22\% were syntactically correct. The wide confidence interval of [$0.0644, 0.1436$] and the mean accuracy of $0.1040$, as shown in Table 1, indicate that LLMs without fine-tuning may not be optimal for generating transforms. The highest accuracy achieved was $0.5728$, using the RandomResizedCrop, ColorJitter, RandomHorizontalFlip, GaussianBlur, ToTensor, and Normalize transforms.

\begin{table}
    \centering
    \fontsize{4}{6}\selectfont % First arg: font size, second: baseline skip
    \begin{tabular}{l c c c}
        \hline
        \textbf{Configuration} & \textbf{Mean Accuracy} & \textbf{Best Accuracy} & \textbf{95\% Confidence Interval} \\
        \hline
        LLM generated(without fine-tuning) & 0.1040 & 0.5728 & [0.0518, 0.1563] \\
        1 transform selected & 0.5256 & \textbf{0.6124} & [0.5234, 0.5279] \\
        2 transforms selected & 0.4832 & 0.6071 & [0.4801, 0.4863] \\
        3 transforms selected & 0.4401 & 0.5983 & [0.4363, 0.4439] \\
        \hline    
    \end{tabular}
    \caption{Performance metrics for the brute-force generation pipeline. Baseline results for the non-fine-tuned LLM are compared against systematic permutations of multiple torchvision transforms. These results provided the performance-labeled metadata required for subsequent iterative supervised fine-tuning.}
    \label{tab:accents}
\end{table}

The brute-force approach generated 6,000 transforms. This method achieved a maximum accuracy of $0.6124$ using the RandomPosterize, Resize, ToTensor, and Normalize transforms. Data transformations with a single selected transform generally outperformed those utilizing two or three transforms, as indicated in Figure 2 and Table 1, which show an increased confidence interval with a higher number of selected transforms. Several data transformations that outperformed the best-performing transform in the LEMUR dataset, when trained with identical hyperparameters and the CIFAR-10 dataset using ResNet, were incorporated into the LEMUR dataset.
\begin{figure}[h]
    \centering
		\includegraphics[width=1.0\linewidth]{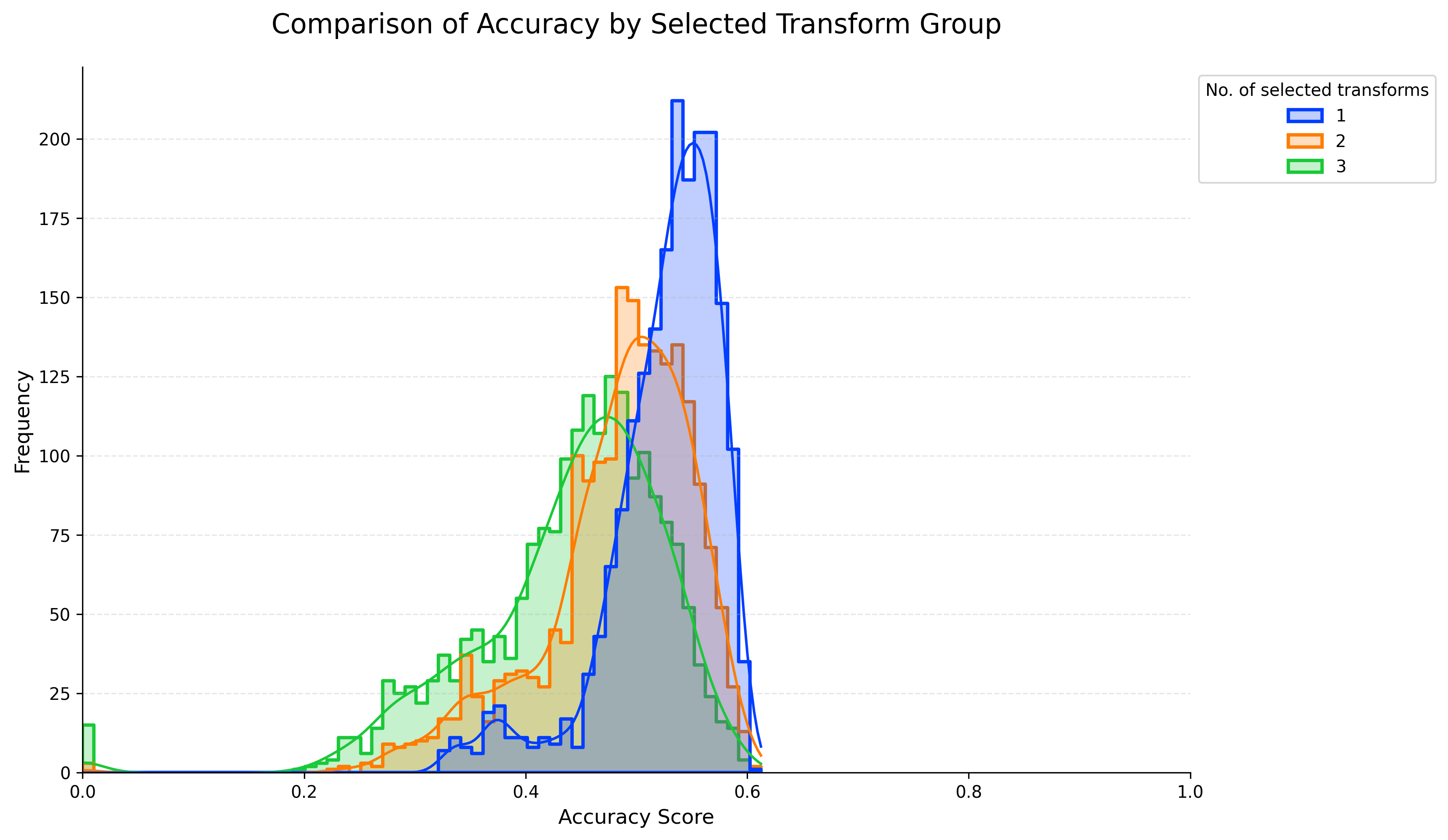}
		\caption{Accuracy distribution of data transformation functions grouped by the number of selected transforms. Single-transform configurations exhibit higher mean accuracy and lower variance compared to compositions of multiple transforms.}
		\label{fig:}	
	\end{figure}
    
The fine-tuning process began with a curated dataset comprising 2,361 pairs of transforms and evaluations obtained through a brute-force approach. Data redundancy was reduced by removing duplicate transform files. Additionally, 1,180 augmented samples were added. Each sample had the input Resize parameter explicitly set to 256. After this initial configuration, the dataset was dynamically expanded by incorporating any new LLM-generated transform with an accuracy greater than 0.55 into the training set for subsequent iterations. The performance of the generated data transform functions was tracked over 28 fine-tuning epochs (A0 to A27), with 10 transforms generated per epoch. Figure 3 shows the LLM's performance during the fine-tuning loop which includes the mean accuracy of all valid transformations and the maximum accuracy achieved by the best single transformation per epoch.

\begin{figure}[h]
    \centering
		\includegraphics[width=1.0\linewidth]{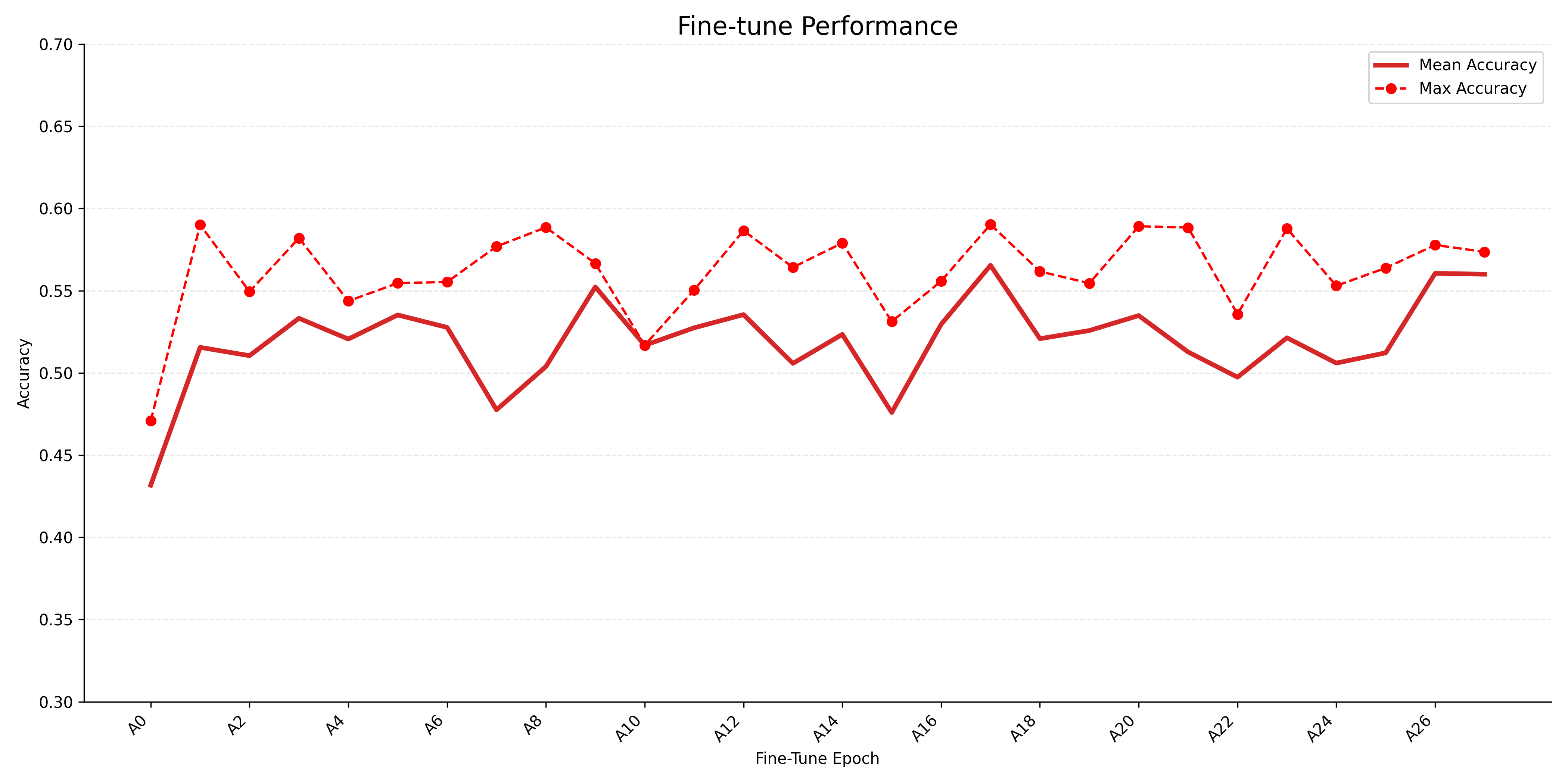}
		\caption{Mean and maximum accuracy of generated transformations across fine-tuning epochs. The mean accuracy exhibits a steadily upward trajectory (r = 0.34) over successive epochs, indicating improved overall generation quality, while the maximum accuracy remains relatively stable, suggesting consistent rediscovery of high-performing transformations.}
		\label{fig:}	
	\end{figure}
    
The mean accuracy increased from $0.4317$ at epoch A0 to $0.56$ at epoch A27. This improvement demonstrates that the fine-tuning process effectively aligned the LLM weights with high-performing transform code. The model generated candidates that led to better convergence after exposure to more positive examples. Maximum accuracy did not show a clear monotonic increase but remained stable. It indicates the model reliably rediscovered or slightly improved the best-known solutions. The gap between mean and maximum accuracy narrowed in final epochs, showing reduced variance. The model generated more consistently effective transformations and fewer low-quality outliers as shown in Figure 3. Qualitative code analysis confirmed that the model learned the inductive bias from the augmented dataset. The generator produced transforms with various resolution parameters, such as 224, 256, and 32. This result shows that the model developed a semantic relation between parameters like Resize(256) and high-accuracy rewards, rather than memorizing syntax.

\begin{figure}[h]
 \centering		
 \includegraphics[width=1.0\linewidth]{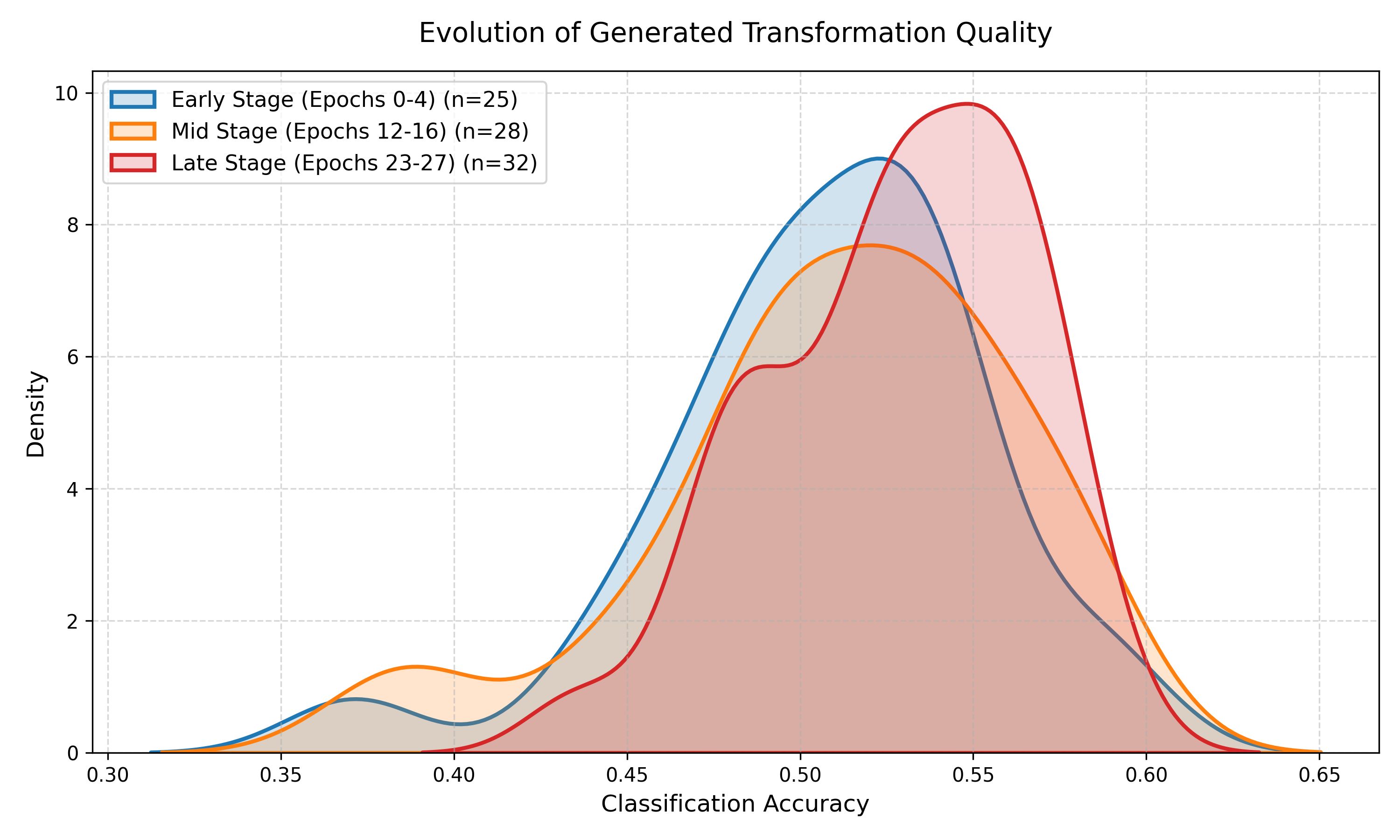}
		\caption{Evolution of Generated Transformation Quality. Kernel Density Estimation (KDE) of validation accuracy across fine-tuning stages. The shift from Early (blue) to Late (red) epochs indicates the model effectively minimizes the generation of low-performing code and converges on a high-performance semantic region}
		\label{fig:descarteslogo}
\end{figure}

\subsection{Efficiency of Fine-Tuning vs. Brute Force}
    
Although high-performing transforms were successfully found using the brute-force search, 6000 candidates had to be generated and evaluated in order to identify them. In contrast, the Fine-tuned LLM showed better sample efficiency. Despite generating only 10 candidates per epoch (280 candidates in total), the model consistently produced transformations with accuracies over the $0.55$ threshold. The fine-tuning process effectively "distilled" the knowledge from the brute-force dataset into the model weights. The LLM did not need to explore thousands of random possibilities. Instead, it quickly converged on the optimal strategy. The refined model could provide competitive augmentations at a significantly higher rate than the random brute-force search. 

\begin{figure}[h]
    \centering
		\includegraphics[width=1.0\linewidth]{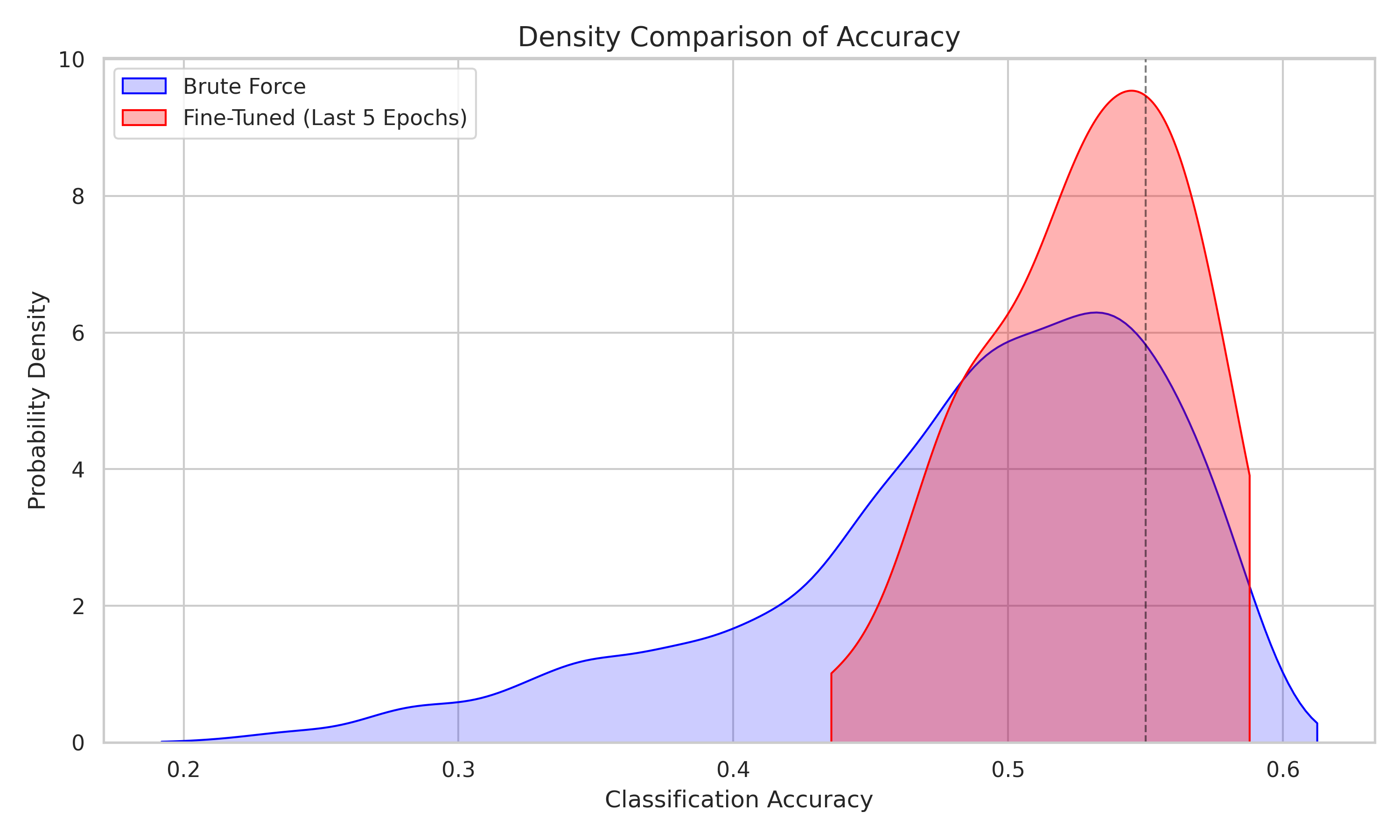}
		\caption{Impact of performance-aware SFT on transformation efficacy. Comparison of classification accuracies achieved by transforms generated at different stages of the iterative loop. The narrowing of the variance and the upward shift in the median accuracy provide empirical evidence that the LLM is successfully internalizing the semantic performance cues from the metadata repository.}
		\label{fig:}	
	\end{figure}
    
\section{Ablation Study}

\subsection*{Effect of Redundancy in Dataset}
    
We used a dataset of 6,000 pairs from the brute force approach, along with transforms generated from previous fine-tuning iterations to understand how data volume and redundancy affect the model's generative stability and convergence. Transforms files containing errors, such as invalid syntax, were also included in the later fine-tuning iterations and assigned an accuracy of $0.0$. There was also significant redundancy, with multiple files containing identical transformation logic differing only by random seed values.

\begin{figure}[h]
    \centering
		\includegraphics[width=1.0\linewidth]{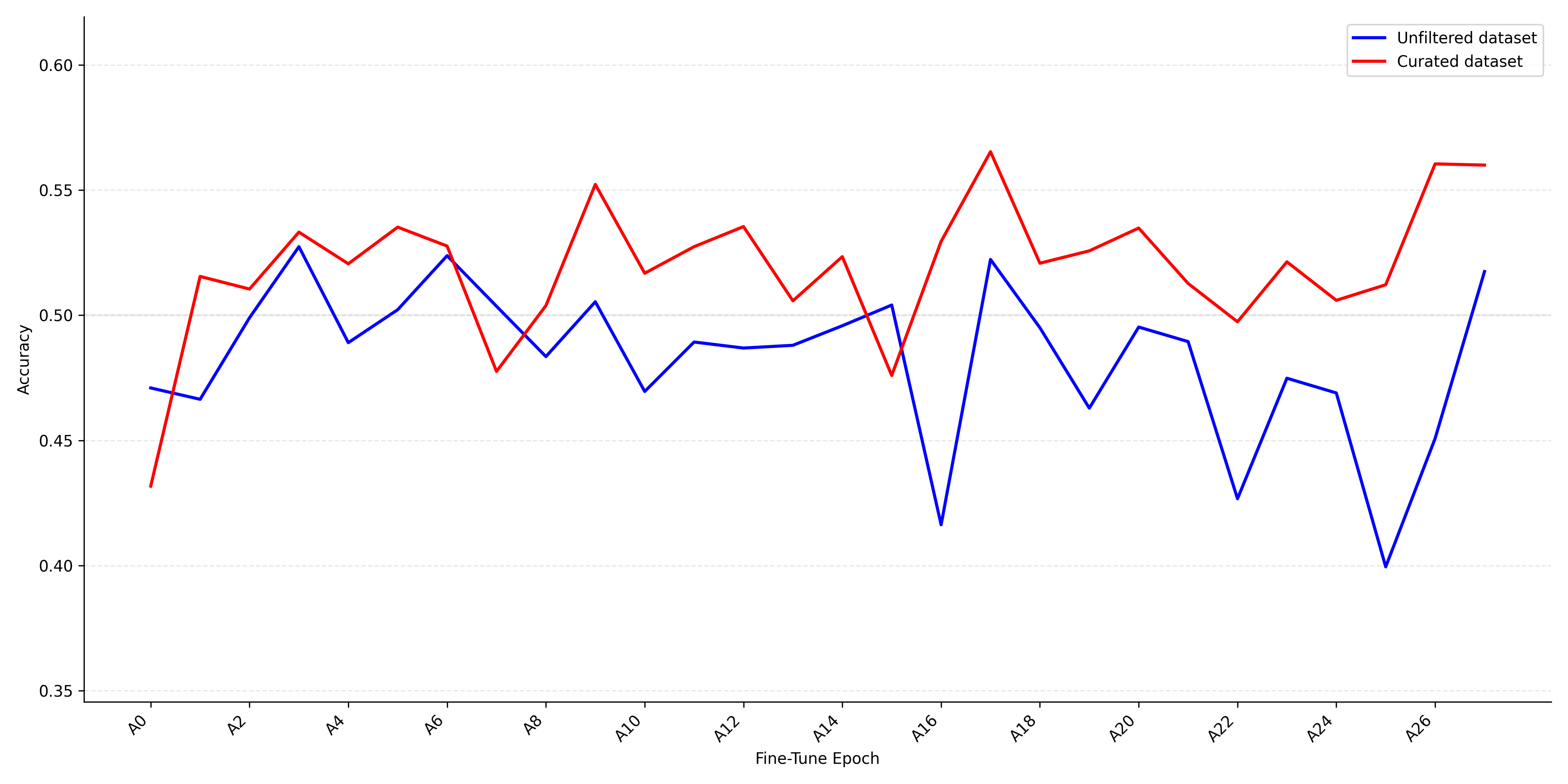}
		\caption{Impact of dataset composition on fine-tuning performance. Mean validation accuracy across fine-tuning epochs for curated and unfiltered datasets. The curated dataset yields more stable convergence and higher accuracy, while redundancy and noisy samples in the unfiltered dataset hinder semantic learning.}
		\label{fig:}	
	\end{figure}
    
As shown in Figure 6, the curated dataset showed a positive trajectory, consistently outperforming fine-tuning with the Unfiltered dataset and achieving mean accuracies exceeding $0.56$. Although the large volume of the dataset enabled the model to generate valid Python syntax, the duplicate files hindered the optimization. The model likely "memorized" frequent file patterns rather than learning to distinguish the specific semantic features that contribute to higher accuracy.

\subsection*{Effect of Prompt Engineering}

To evaluate how the model responds to different instruction formats, we compared two approaches. Firstly, the Direct approach used a simple prompt focused on code generation. Then Structured CoT (Chain-of-Thought) with Constraints used a verbose prompt requiring analysis before code, along with explicit negative constraints.

\begin{figure}[h]
    \centering
		\includegraphics[width=1.0\linewidth]{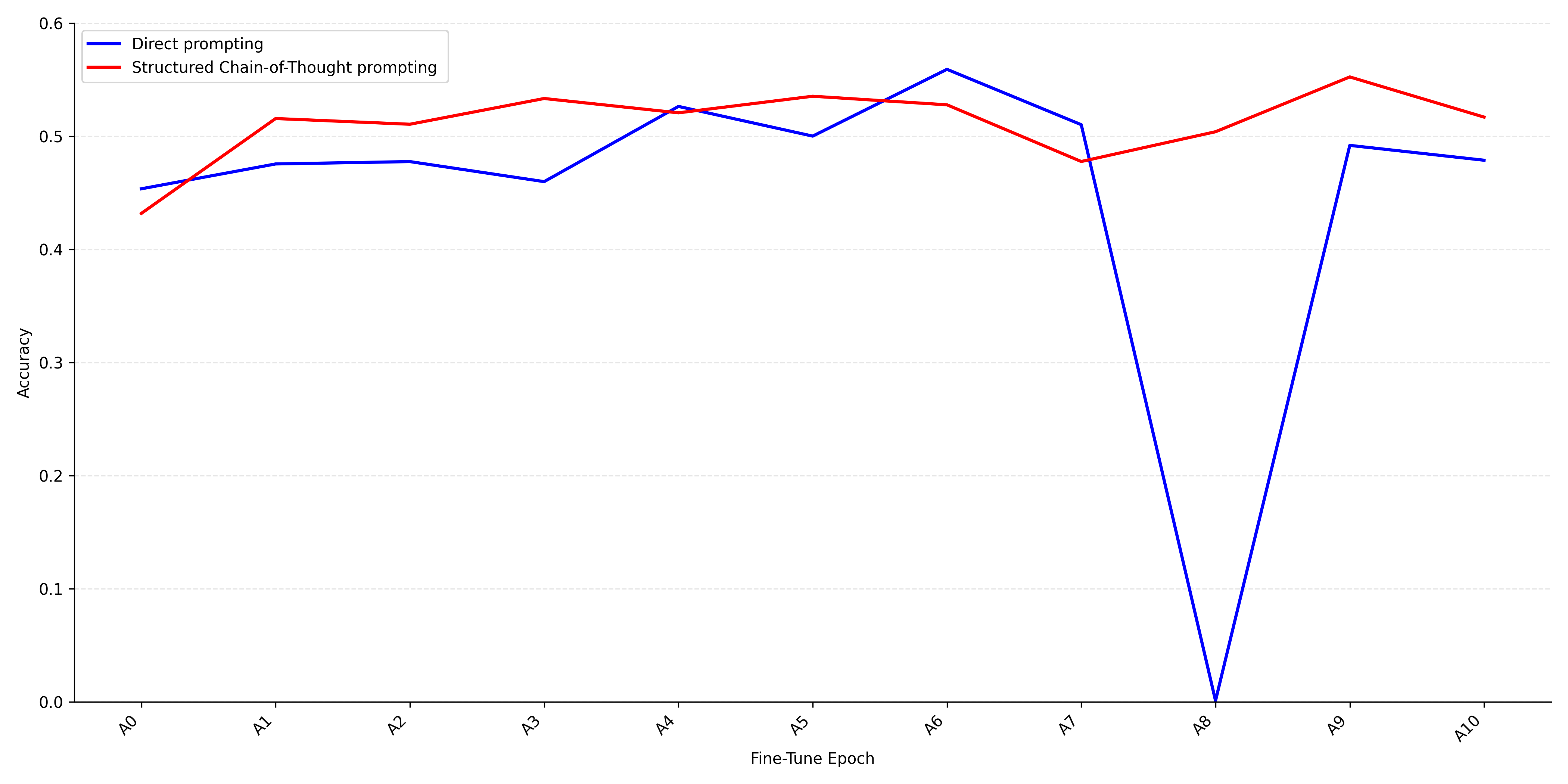}
		\caption{Comparative performance of Direct vs. Structured Chain-of-Thought prompting. The results indicate that Structured CoT is prone to optimization instability, evidenced by the total performance drop at epoch 8. The Direct approach remains more stable and effective, successfully internalizing semantic cues without the noise introduced by verbose reasoning requirements.}
		\label{fig:}	
	\end{figure}

As shown in Figure 7, the Direct approach demonstrated better stability and convergence than the Structured approach. Although the mean accuracy peaked at epoch A6, it dropped below $0.48$ by epoch A10. This drop suggests that the extra tokens needed for analysis introduced noise. The model struggled to balance generating clear reasoning with valid Python syntax.

The Direct approach shows a consistent positive trend comparatively, peaking above $0.55$ at epoch A10. Removing unnecessary instructions allowed the fine-tuning signal to focus on the target code. Moreover, removing the "Negative Constraints" and reasoning requirements concentrated the model's attention solely on the target output (the transform code). The inclusion of "Negative Constraints" may have mistakenly directed the model's focus towards the artifacts (SVG, HTML) that it was intended to avoid, or diluted the context window with unnecessary instructions.

\section{Conclusion}

This work presents a performance-aware, closed-loop framework for the autonomous synthesis of data transformations, demonstrating that large language models can be effectively grounded in empirical training outcomes. By constructing a novel repository of over 6{,}000 empirically evaluated PyTorch augmentation functions and fine-tuning via Low-Rank Adaptation, we induce task-level reasoning in the generator without relying on explicit symbolic rewards, reinforcement learning, or handcrafted objectives.

Our experiments yield several insights relevant to the design of future automated code synthesis and learning systems. First, empirical alignment through iterative fine-tuning shifts the generative distribution from random code synthesis toward informed, task-aligned design, achieving up to a 600$\times$ reduction in evaluated candidates compared to brute-force discovery while improving mean accuracy from $0.43$ to $0.56$ and preserving competitive peak performance. Second, qualitative and quantitative analyses show that the model internalizes semantic performance cues—such as the benefits of resolution scaling (e.g., \texttt{Resize(256)})—rather than memorizing transformation syntax, indicating meaningful generalization beyond surface-level patterns. Third, ablation studies reveal a critical trade-off between prompt complexity and optimization stability: while direct prompting supports reliable improvement, structured Chain-of-Thought prompting introduces syntactic instability that leads to catastrophic performance degradation, underscoring the fragility of complex reasoning formats in performance-critical code-generation tasks.

Taken together, these results demonstrate that grounding LLMs in non-textual, empirical feedback loops provides a robust alternative to symbolic or reward-based alignment for complex downstream objectives. By addressing the limited diversity and supervision of existing augmentation datasets, this work lays a scalable foundation for autonomous machine learning pipelines. Future work will focus on improving syntactic robustness at larger model scales and extending this grounded reasoning framework to multimodal data, broader architectural families, and more heterogeneous optimization tasks.

\section{Limitations}
We limited our experimental setup, which includes both the generative fine-tuning loop and evaluation, to the ResNet architecture and a single dataset. Furthermore, the augmentations generated are implicitly specialized for this specific configuration. We assume that the best augmentation strategy depends on the interactions between model architecture and data distribution rather than a single transformation function that works well in every situation. Since we did not extend the fine-tuning process to alternative architectures or datasets, we cannot quantify the LLM’s ability to dynamically adapt its generation to discover the "best fit" solutions. Furthermore, the constraint to a single-epoch training limits our assessment of the long-term convergence stability of the neural network.

\bibliography{custom}

\appendix

%\section{Example Appendix}
%\label{sec:appendix}

% This is an appendix.

\end{document}